\documentclass{article}

\usepackage{graphicx}
\usepackage{subcaption}
\usepackage{multirow}
\usepackage{booktabs} 

\usepackage[utf8]{inputenc} 
\usepackage[T1]{fontenc}    
\usepackage{hyperref}       
\usepackage{url}            
\usepackage{booktabs}       
\usepackage{amsfonts}       
\usepackage{nicefrac}       
\usepackage{microtype}      

\usepackage{hyperref}
\usepackage[capitalise]{cleveref}
\usepackage{multirow}
\usepackage{tikz}
\usepackage{amsmath}
\usepackage{amssymb}
\usepackage{wrapfig}
\usepackage{floatrow}

\usepackage[square,sort,comma,numbers]{natbib}

\usetikzlibrary{arrows,positioning,shadows.blur,shapes.geometric,patterns, decorations.pathreplacing}

\usepackage[final,nonatbib]{neurips_2019}

\title{Hierarchical Decision Making by Generating and Following Natural Language Instructions}

\author{
  Hengyuan Hu\thanks{Equal Contribution.} \\
  Facebook AI Research \\
  \texttt{hengyuan@fb.com} \\
  \And
  Denis Yarats\footnotemark[1]\\
  New York University \& Facebook AI Research \\
  \texttt{denisyarats@cs.nyu.edu} \\
  \And
  Qucheng Gong \\
  Facebook AI Research \\
  \texttt{qucheng@fb.com} \\
  \And
  Yuandong Tian \\
  Facebook AI Research \\
  \texttt{yuandong@fb.com} \\
  \And
  Mike Lewis \\
  Facebook AI Research  \\
  \texttt{mikelewis@fb.com}
}

\begin{document}

\maketitle

\begin{abstract}
We explore using latent natural language instructions as an expressive and compositional representation of complex actions for hierarchical decision making.
Rather than directly selecting micro-actions, our agent first generates a latent plan in natural language, which is then executed by a separate model.
We introduce a challenging real-time strategy game environment in which the actions of a large number of units must be coordinated across long time scales. 
We gather a dataset of 76 thousand pairs of instructions and executions from human play, and train \emph{instructor} and \emph{executor} models.
Experiments show that models using natural language as a latent variable significantly outperform models that directly imitate human actions.
The compositional structure of language proves crucial to its effectiveness for action representation.
We also release our code, models and data\footnote{A demo is available at www.minirts.net}\footnote{Our code is open-sourced at www.github.com/facebookresearch/minirts}.
\end{abstract}

\newcommand{\execrnn}[0]{\textsc{Rnn}}
\newcommand{\execbow}[0]{\textsc{BoW}}
\newcommand{\execrand}[0]{\textsc{ExecutorOnly}}
\newcommand{\execonehot}[0]{\textsc{OneHot}}

\section{Introduction}


Many complex problems can be naturally decomposed into steps of high level planning and low level control.
However, plan representation is challenging---manually specifying macro-actions requires significant domain expertise, limiting generality and scalability \citep{sutton1999,tessler2016}, but learning composite actions from only end-task supervision can result in the hierarchy collapsing to a single action \citep{bacon2016}.

\begin{figure*}[t!]

\centering
\includegraphics[width=0.7\linewidth]{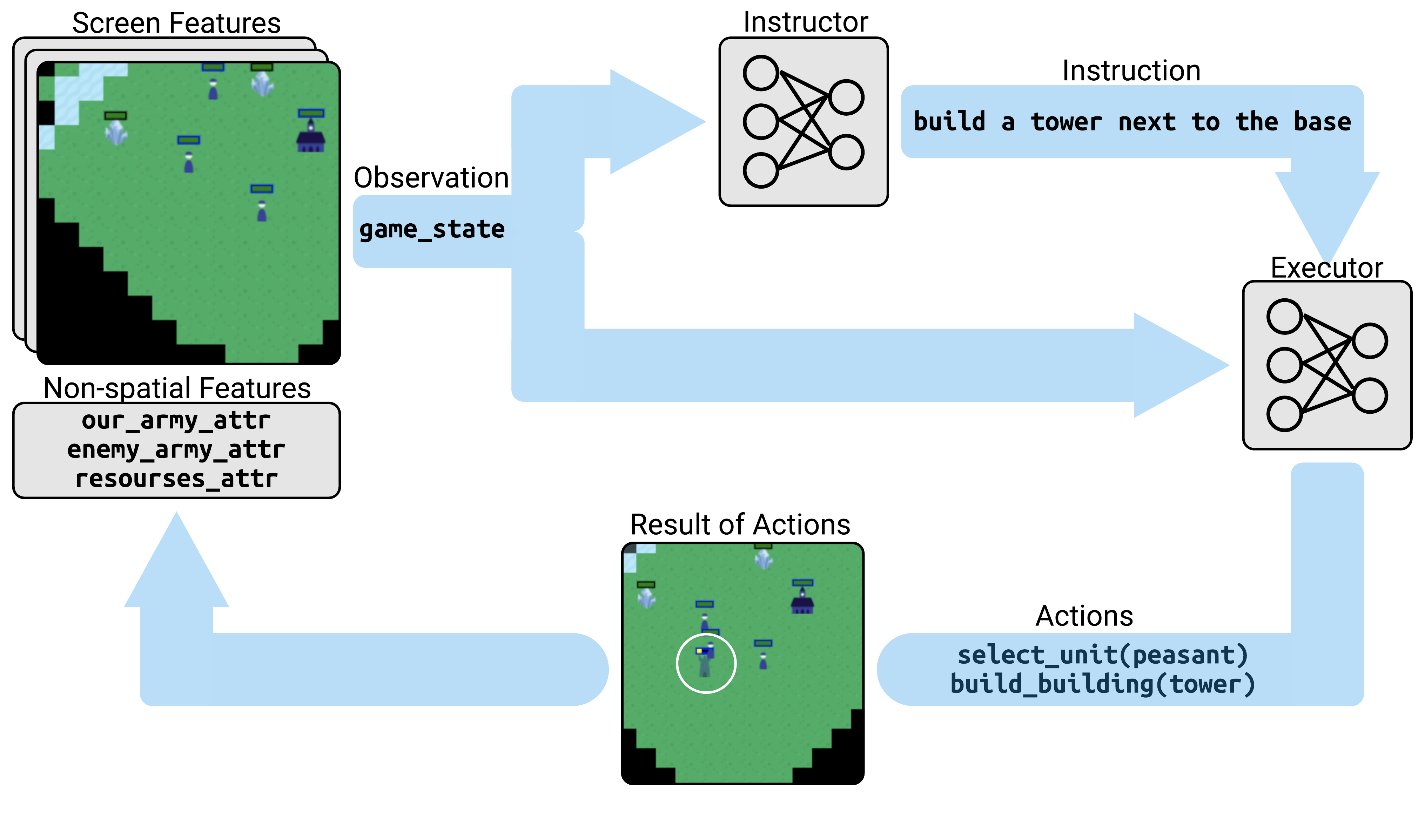}

\caption{\label{figure:game_graph}
Two agents, designated \emph{instructor} and \emph{executor} collaboratively play a real-time strategy game (\S\ref{section:game}).
The \emph{instructor} iteratively formulates plans and issues instructions in natural language to the \emph{executor}, who then executes them as a sequence of actions.
We first gather a dataset of humans playing each role (\S\ref{section:dataset}).
We then train models to imitate humans actions in each role (\S\ref{section:models}).
}
\end{figure*}

We explore representing complex actions as natural language instructions.
Language can express arbitrary goals, and has compositional structure that allows generalization across commands \citep{oh2017zero, andreas2017}. 
Our agent has a two-level hierarchy, where a high-level \emph{instructor} model communicates a sub-goal in natural language to a low-level \emph{executor} model, which then interacts with the environment (\cref{figure:game_graph}). Both models are trained to imitate humans playing the roles. 
This approach decomposes decision making into planning and execution modules, with a natural language interface between them.


We gather example instructions and executions from two humans collaborating in a complex game.
Both players have access to the same partial information about the game state.
One player acts as the \emph{instructor}, and periodically issues instructions to the other player (the \emph{executor}), but has no direct control on the environment. The \emph{executor}
acts to complete the \emph{instruction}. This setup forces the \emph{instructor} to focus on high-level planning, while the \emph{executor} concentrates on low-level control. 

To test our approach, we introduce a real-time strategy (RTS) game, developing an environment based on \citep{tian2017elf}. A key property of our game is the \emph{rock-paper-scissors} unit attack dynamic, which emphasises strategic planning over micro control. Our game environment is a challenging decision making task, because of exponentially large state-action spaces, partial observability, and the variety of effective strategies. However, it is relatively intuitive for humans, easing data collection. 

Using this framework, we gather a dataset of 5392 games, where two humans (the \emph{instructor} and \emph{executor}) control an agent against rule-based opponents. The dataset contains 76 thousand pairs of human instructions and executions, spanning a wide range of strategies. 
This dataset poses challenges for both instruction generation and execution, as instructions may apply to different subsets of units, and multiple instructions may apply at a given time. 
We design models for both problems, and extensive experiments show that using latent language significantly improves performance.

In summary, we introduce a challenging RTS environment for sequential decision making, and a corresponding dataset of instruction-execution mappings. We develop novel model architectures with planning and control components, connected with a natural language interface. 
Our agent with latent language significantly outperforms agents that directly imitate human actions, and we show that exploiting the compositional structure of language improves performance by allowing generalization across a large instruction set.
We also release our code, models and data.

\section{Task Environment}
\label{section:game}
We implement our approach for an RTS game, which has several attractive properties compared to traditional reinforcement learning environments, such as Atari~\citep{mnih2013} or grid worlds~\citep{sukhbaatar15}. The large state and action spaces mean that planning at different levels of abstraction is beneficial for both humans and machines. However, manually designed macro-actions typically do not match strong human performance, because of the unbounded space of possible strategies~\citep{synnaeve16,vinyals2017}. Even with simple rules, adversarial games have the scope for complex emergent behaviour. 

We introduce a new RTS game environment, which distills the key features of more complex games while being faster to simulate and more tractable to learn. Current RTS environments, such as StarCraft, have dozens of unit types, adding large overheads for new players to learn the game. Our new  environment is based on MiniRTS~\citep{tian2017elf}.
It has a set of 7 unit types, designed with a \emph{rock-paper-scissors} dynamic such that each has some units it is effective against and vulnerable to.
Maps are randomly generated each game to force models to adapt to their environment as well as their opponent.
The game is designed to be intuitive for new players (for example, catapults have long range and are effective against buildings).
Numerous strategies are viable, and the game presents players with dilemmas such as whether to attack early or focus on resource gathering, or whether to commit to a strategy or to attempt to scout for the opponent's strategy first. Overall, the game is easy for humans to learn, but challenging for machines due to the large action space, imperfect information, and need to adapt strategies to both the map and opponent. See the Appendix for more details.

\section{Dataset}

\label{section:dataset}
To learn to describe actions with natural language, we gather a dataset of two humans playing collaboratively against a rule-based opponent.  Both players have access to the same information about the game state, but have different roles. 
One is designated the \emph{instructor}, and is responsible for designing strategies and describing them in natural language, but has no direct control. The other player, the \emph{executor}, must ground the instructions into low level control. The \emph{executor}'s goal is to carry out commands, not to try to win the game. 
This setup causes humans to focus on either planning or control, and provides supervision for both generating and executing instructions. 

\begin{wraptable}{r}{0.49\linewidth}
\centering
\begin{tabular}{|l|c|}
\hline
Statistic & Value \\
\hline
Total games                             & 5392 \\
Win rate                                    & 58.6\% \\
Total instructions                     & 76045\\
Unique instructions            & 50669 \\
Total words                            & 483650 \\
Unique words                      & 5007 \\
\# words per instruction     & 9.54 \\
\# instructions per game     & 14.1 \\
\hline
\end{tabular}
\caption{\label{table:dataset_stats}
We gather a large language dataset for instruction generation and following. Major challenges include the wide range of unique instructions and the large number of low-level actions required to execute each instruction.
}

\end{wraptable}

We collect 5392 games of human teams against our bots.\footnote{Using ParlAI~\citep{miller2017}} Qualitatively, we observe a wide variety of different strategies.
An average game contains 14 natural language instructions and lasts for 16 minutes. Each instruction corresponds to roughly 7 low-level actions, giving a challenging grounding problem (\cref{table:dataset_stats}). The dataset contains over 76 thousand instructions, most of which are unique, and their executions. The diversity of instructions shows the wide range of useful strategies. The instructions contain a number of challenging linguistic phenomena, particularly in terms of reference to locations and units in the game, which often requires pragmatic inference. Instruction execution is typically highly dependent on context. Our dataset is publicly available. For more details, refer to the Appendix.

Analysing the list of instructions (see Appendix), we see that the head of the distribution is dominated by straightforward commands to perform the most frequent actions. However, samples from the complete instruction list reveal many complex compositional instructions, such as \emph{Send one catapult to attack the northern guard tower [and] send a dragon for protection}. We see examples of challenging quantifiers (\emph{Send all but 1 peasant to mine}), anaphora (\emph{Make 2 more cavalry and send them over with the other ones}), spatial references (\emph{Build a new town hall between the two west minerals patches}) and conditionals (\emph{If attacked retreat south}).

\section{Model}

\label{section:models}
We factorize agent into an \emph{executor} model (\S\ref{section:executor}), which maps instructions and the game states into unit-level actions of the environment, and an \emph{instructor} model (\S\ref{section:instructor}), which generates language instructions given the game states. We train both models with human supervision (\S\ref{section:sv}). 


\subsection{Game Observation Encoder}
\label{section:encoder}
\begin{figure}[!t]
\centering
\includegraphics[width=\linewidth]{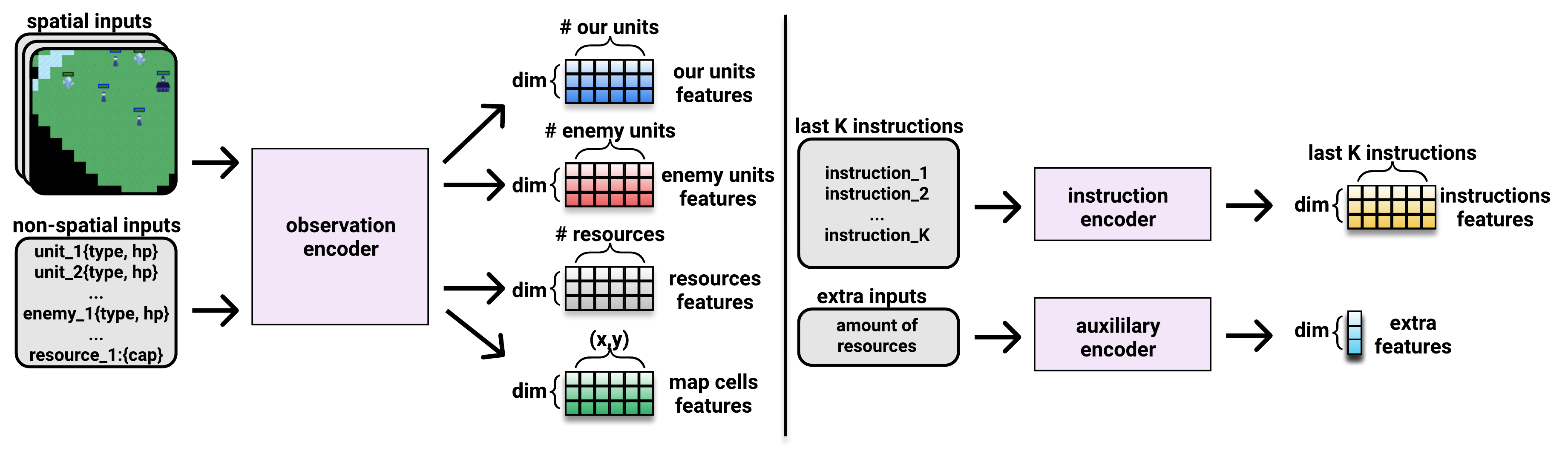}

\caption{\label{figure:encoder_model}
At each time step of the environment we encode spatial observations (e.g. the game map) and non-spatial internal states for each game object (e.g. units, buildings, or resources) via the observation encoder, which produces  separate feature vectors for each unit, resource, or discrete map locations. We also embed each of the last $K$ natural language instructions into individual instruction feature vectors. Lastly, we learn features for all the other global game attributes by employing the auxiliary encoder. We then use these features for both the \emph{executor} and \emph{instructor} networks.
 }
\end{figure}

We condition both the \emph{instructor} and \emph{executor} models on a fixed-sized representation of the current game state, which we construct from a spatial map observation, internal states of visible units, and several previous natural language instructions. (\cref{figure:encoder_model}). We detail each individual encoder below.

\subsubsection{Spatial Inputs Encoder}
\label{section:game_state_encoder}
We encode the spatial information of the game map using a convolutional network. We discretize the map into a $32\times32$ grid and extract different bits of information from it using separate channels. For example, three of those channels provide binary indication of a particular cell visibility, which indicates \textsc{Invisible}, \textsc{Seen}, and \textsc{Visible} states. We also have a separate channel per unit type to record the number of units in each spatial position for both our and enemy units separately.  Note that due to "fog-of-war", not all enemy units are visible to the player.  See the Appendix for more details.

We apply several $3\times3$ convolutional layers that preserve the spatial dimensions to the input tensor. Then we use 4 sets of different weights to project the shared convolutional features onto different 2D features spaces, namely \textsc{Our Units}, \textsc{Enemy Units}, \textsc{Resources}, and \textsc{Map Cells}. We then use  $(x,y)$ locations for units, resources, or map cells to extract their features vectors from corresponding 2D features spaces.

\subsubsection{Non-spatial Inputs Encoder}
We also take advantage of non-spatial attributes and internal state for game objects. Specifically, we improve features vectors for \textsc{Our Units} and \textsc{Enemy Units} by adding encodings of units health points, previous and current actions. If an enemy unit goes out the players visibility, we respect this by using the state of the unit's attributes from the last moment we saw it.
We project attribute features onto the same dimensionality of the spatial features and do a element-wise multiplication to get the final set of \textsc{Our Units} and \textsc{Enemy Units} features vectors.


\subsubsection{Instruction Encoders}
\label{section:instruction_encoder}

The state also contains a fixed-size representation of the current instruction. We experiment with:
\begin{itemize}
    \item An instruction-independent model (\execrand), that directly mimics human actions.
    \item A non-compositional encoder (\execonehot) which embeds each instruction with no parameter sharing across instructions (rare instructions are represented with an \emph{unknown} embedding).
    \item A bag-of-words encoder (\execbow), where an instruction encoding is a sum of word embeddings. This model tests if the compositionality of language improves generalization.
    \item An RNN encoder (\execrnn), which is order-aware. Unlike \execbow, this approach can differentiate instructions such as \emph{attack the dragon with the archer} and \emph{attack the archer with the dragon}.
\end{itemize}


\subsubsection{Auxiliary Encoder}
Finally, we encode additional game context, such as the amount of money the player has, through a simple MLP to get the \textsc{Extra} features vector.

\subsection{Executor Model}
\label{section:executor}

\begin{wrapfigure}{rh}{0.5\linewidth}
\centering
\includegraphics[width=0.9\linewidth]{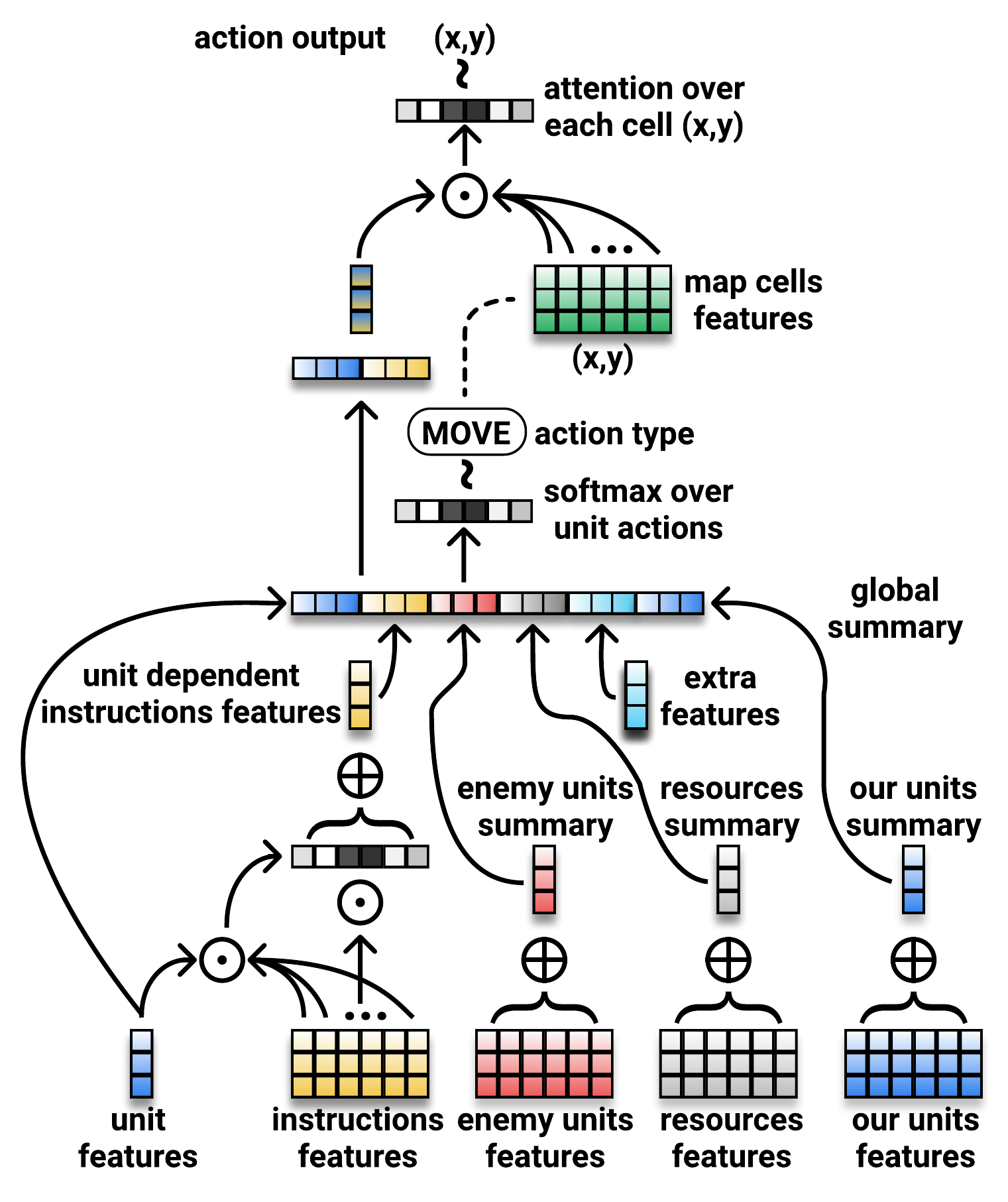}

\caption{\label{figure:executor_model}
Modeling an action for an unit requires predicting an action type based on the \textbf{global summary} of current observation, and then, depending on the predicted action type, computing a probability distribution over a set of the action targets. In this case, the \textsc{Move} action is sampled, which uses the map cells features as the action targets. 
}
\end{wrapfigure}

The \emph{executor} predicts an action for every unit controlled by the agent based on the global  summary of the current observation. We predict an action for each of the player's units by choosing over an \textsc{Action Type} first, and then selecting the \textsc{Action Output}. There are 7 action types available: \textsc{Idle, Continue, Gather, Attack, Train Unit, Build Building, Move}. \textsc{Action Output} specifies the target output for the action, such as a target location for the \textsc{Move} action, or the unit type for \textsc{Train Unit}. \cref{figure:executor_model} gives an overview of the \emph{executor} design, also refer to the Appendix.


For each unit, we consider a history of recent $N$ instructions ($N=5$ in all our experiments), because some units may still be focusing on a previous instruction that has long term effect like \emph{keep scouting} or \emph{build 3 peasants}. To encode the $N$ instructions, we first embed them in isolation with the \ref{section:instruction_encoder}. We take $K$ that represents how many frames have passed since that instruction gets issued and compute $H = \max(H_{max}, K / B)$ where $H_{max}, B$ are constants defining the number of bins and bin size. We also take $O = 1, 2, ..., N$ that represents the temporal ordering of those instructions. We embed $O$ and $H$ and concatenate the embeddings with language embedding. Dot product attention is used to compute an attention score between a unit and recent instructions and then a unit dependent instruction representation is obtained through a weighted sum of history instruction embeddings using attention score as weight.

We use the same observation encoder (\S\ref{section:encoder}) to obtain the features mentioned above. To form a global summary, we sum our unit features, enemy unit features, and resource features respectively and then concatenate together with \textsc{Extra} features. 

To decide the action for each unit, we first feed the concatenation of the unit feature, unit depending instruction feature and the global summary into a multi-layer neural classifier to sample an \textsc{Action Type}. Depending on the action type, we then feed inputs into different action-specific classifiers to sample \textsc{Action Output}. 
In the action argument classifier, the unit is represented by the concatenation of unit feature and instruction feature, and the targets are represented by different target embeddings. For \textsc{Attack}, the target embeddings are enemy features; for \textsc{Gather}; the target embeddings are resource features; for \textsc{Move}, the target embeddings are map features; for \textsc{Train Unit}, the target embeddings are embeddings of unit types; for \textsc{Build Building}, the target embeddings are embeddings of unit types and map features, and we sample type and location independently. The distribution over targets for the action is computed by taking the dot product between the unit representation and each target, followed by a softmax.

We add an additional binary classifier, \textsc{Global Continue}, that takes the global summary and \textbf{current} instruction embedding as an input to predict whether all the agent's units should continue working on their previous action.

\subsection{Instructor Model}

\label{section:instructor_model}
The \emph{instructor} maps the game state to instructions. It uses the game observation encoder (\S\ref{section:encoder}) to compute a global summary and \textbf{current} instruction embedding similar to the \emph{executor}. We experiment with two model types:

\paragraph{Discriminative Models} These models operate on a fixed set of instructions. Each instruction is encoded as a fixed-size vector, and the dot product of this encoding and the game state encoding is fed into a softmax classifier over the set of instructions. As in \S\ref{section:instruction_encoder}, we consider non-compositional (\execonehot), bag-of-words (\execbow) and \textsc{Rnn Discriminative} encoders.

\paragraph{Generative Model} The discriminative models can only choose between a fixed set of instructions. We also train a generative model, \textsc{Rnn Generative}, which generates instructions autoregressively. To compare likelihoods with the discriminative models, which consider a fixed set of instructions, we re-normalize the probability of an instruction over the space of instructions in the set.


The \emph{instructor} model must also decide at each time-step whether to issue a new command, or leave the \emph{executor} to follow previous instructions. We add a simple binary classifier that conditions on the global feature, and only sample a new instruction if the result is positive.

Because the game is only partially observable, it is important to consider historical information when deciding an instruction. 
For simplicity, we add a running average of the number of enemy units of each type that have appeared in the visible region as an extra input to the model.
To make the \emph{instructor} model aware of how long the current instruction has been executed, we add an extra input representing number of time-step passed since the issuance of current instruction. 
As mentioned above, these extra inputs are fed into separate MLPs and become part of the \textsc{Extra} feature.


\label{section:instructor}

\subsection{Training}

\label{section:training}

\label{section:sv}


Since one game may last for tens of thousands of frames, it is not feasible nor necessary to use all frames for training. Instead, we take one frame every $K$ frames to form the supervised learning dataset. To preserve unit level actions for the \emph{executor} training, we put all actions that happen in $[tK, (t+1)K)$ frames onto the $tK$th frame if possible. For actions that cannot happen on the $tK$th frame, such as actions for new units built after $tK$th frame, we simply discard them. 

Humans players sometimes did not execute instructions immediately. To ensure our \emph{executor} acts promptly, we filter out action-less frames between a new instruction and first new actions.

\subsubsection{Executor Model}
The \emph{executor} is trained to minimize the following negative log-likelihood loss:
\begin{align*}
\mathcal{L} &= -\log P_{\mathrm{cont}}(c | s) - (1 - c) \cdot \sum_{i=1}^{|u|} \log P_{\mathrm{A}}(a_{u_i} | s)
\end{align*}
where $s$ represents game state and instruction, $P_{\mathrm{cont}}(\cdot|s)$ is the \emph{executor} \textsc{Global Continue} classifier (see \S\ref{section:executor}), $c$ is a binary label that is 1 if all units should continue their previous action, $P_{A}(a_{u_i} | s)$ is the likelihood of unit $i$ doing the correct action $a_{u_i}$.

\subsubsection{Instructor Model}
The loss for the \emph{instructor} model is the sum of a loss for deciding whether to issue a new instruction, and the loss for issuing the correct instruction:
\begin{align*}
    \mathcal{L} &= -\log P_{\mathrm{cont}}(c | s) -  (1 - c)\cdot \mathcal{L}_{\mathrm{lang}}
\end{align*}
where $s$ represents game state and current instruction, $P_{\mathrm{cont}}(\cdot|s)$ is the continue classifier, and $c$ is a binary label with $c = 1$ indicating that no new instruction is issued. 
The language loss $\mathcal{L}_{\mathrm{lang}}$ is the loss for choosing the correct instruction, and is defined separately for each model.

For \textsc{OneHot} instructor, $\mathcal{L}_{\mathrm{lang}}$ is simply negative log-likelihood of a categorical classifier over a pool of $N$ instructions. If the true target is not in the candidate pool $\mathcal{L}_{\mathrm{lang}}$ is 0.

Because \textsc{BoW} and \textsc{Rnn Discriminative} can compositionally encode any instruction (in contrast to \textsc{OneHot}), we can additionally train on instructions from outside the candidate pool. 
To do this, we encode the true instruction, and discriminate against the $N$ instructions in the candidate pool and another $M$ randomly sampled instructions. The true target is forced to appear in the $M+N$ candidates. We then use the NLL of the true target as language loss. This approach approximates the expensive softmax over all 40K unique instructions.

For \textsc{Rnn Generative}, the language loss is the standard autoregressive loss.

\section{Experiments}

\begin{table}[t]
\centering
\resizebox{\columnwidth}{!}{
\begin{tabular}{|p{.3\columnwidth}|c|c|}
\hline
Executor Model & Negative Log Likelihood & Win/Lose/Draw Rate (\%) \\
\hline
\execrand    & 3.15 $\pm$ 0.0024 &  41.2/40.7/18.1 \\
\execonehot  & 3.05 $\pm$ 0.0015 &  49.6/37.9/12.5  \\
\execbow     & 2.89 $\pm$ 0.0028 &  54.2/33.9/11.9  \\
\execrnn & \textbf{2.88 $\pm$ 0.0006} & \textbf{57.9/30.5/11.7}  \\
\hline
\end{tabular}
}
\caption{\label{table:executor_loss} Negative log-likelihoods of human actions for \emph{executor} models, and win-rates against \execrand, which does not use latent language. We use the \textsc{Rnn Discriminative} \emph{instructor} with 500 instructions. Modelling instructions compositionally improves performance, showing linguistic structure enables generalization.}

\end{table}


We compare different \emph{executor} (\S\ref{section:executor_eval}) and \emph{instructor} (\S\ref{section:instructor_eval}) models in terms of both likelihoods and end-task performance.
We show that hierarchical models perform better, and that the compositional structure of language improves results by allowing parameter sharing across many instructions.


\begin{table*}[t]
\centering
\resizebox{\columnwidth}{!}{
\begin{tabular}{|l|ccc|ccc|}
\hline
 Instructor Model & \multicolumn{3}{c|}{Negative Log Likelihood} & \multicolumn{3}{c|}{Win/Lose/Draw rate (\%)} \\
(with N instructions) & N=50       & N=250       & N=500      & N=50 & N=250 & N=500 \\ 
\hline
\execonehot                 & 0.662 $\pm$ 0.005 & 0.831 $\pm$ 0.001 & 0.911 $\pm$ 0.005
& 44.6 / 43.4 / 12.0 & 49.7 / 35.9 / 14.3 & 43.1 / 41.1 / 15.7\\
\execbow                    & 0.638 $\pm$ 0.004 & 0.792 $\pm$ 0.001 & 0.869 $\pm$ 0.002 
& 41.3 / 41.2 / 17.5 & 51.5 / 33.3 / 15.3 & 50.5 / 37.1 / 12.5 \\
\textsc{Rnn Discriminative} & \textbf{ 0.618 $\pm$ 0.005} & \textbf{0.764 $\pm$ 0.002} & \textbf{0.826 $\pm$ 0.002}
& 47.8 / 36.5 / 15.7 & 55.4 / 33.1 / 11.5 & \textbf{57.9 / 30.5 / 11.7}  \\
\textsc{Rnn Generative}   &  0.638 $\pm$ 0.006 & 0.794 $\pm$ 0.006 & 0.857 $\pm$ 0.002
&  47.3 / 38.1 / 14.6  & 51.1 / 33.7 / 15.2 & 54.8 / 33.8 / 11.4 \\
\hline
\end{tabular}
}
\caption{\label{table:instructor_result} Win-rates and likelihoods for different \emph{instructor} models, with the $N$ most frequent instructions. Win-rates are against a non-hierarchical \emph{executor} model, and use the \textsc{Rnn} \emph{executor}. Better results are achieved with larger instruction sets and more compositional instruction encoders.}


\end{table*}

\subsection{Executor Model}
\label{section:executor_eval}
The \emph{executor} model learns to ground pairs of states and instructions onto actions.
With over 76 thousand examples, a large action space, and multiple sentences of context, this problem in isolation is one of the largest and most challenging tasks currently available for grounding language in action.

We evaluate \emph{executor} performance with different instruction encoding models (\S\ref{section:instruction_encoder}).
Results are shown in \cref{table:executor_loss}, and show 
that modelling instructions compositionally---by encoding words (\execbow) and word order (\execrnn)---improves both the likelihoods of human actions, and win-rates over non-compositional \emph{instructor} models (\execonehot). The gain increases with larger instruction sets, demonstrating that a wide range of instructions are helpful, and that exploiting the compositional structure of language is crucial for generalization across large instruction sets.

We additionally ablate the importance of considering multiple recent instructions during execution (our model performs attention over the most recent 5 commands \S\ref{section:executor}). When considering only the current instruction with the RNN \emph{executor}, we find performance drops to a win-rate of 52.9 (from 57.9) and negative log likelihood worsens from 2.88 to 2.93.



\subsection{Instructor Model}
\label{section:instructor_eval}
We compare different \emph{instructor} models for mapping game states to instructions. As in \S\ref{section:executor_eval}, we experiment with non-compositional, bag-of-words and RNN models for instruction generation. For the RNNs, we train both a discriminative model (which maps complete instructions onto vectors, and then chooses between them) and a generative model that outputs words auto-regressively. 

Evaluating language generation quality is challenging, as many instructions may be reasonable in a given situation, and they may have little word overlap. We therefore compare the likelihood of the human instructions. Our models choose from a fixed set of instructions, so we measure the likelihood of choosing the correct instruction, normalized over all instructions in the set. 
Likelihoods across different instructions sets are not comparable. 

\cref{table:instructor_result} shows that, as \S\ref{section:executor_eval}, more structured instruction models give better likelihoods---particularly for larger instruction sets, which are harder to model non-compositionally. 


We compare the win-rate of our models against a baseline which directly imitates human actions (without latent language). All latent instruction models outperform this baseline. More compositional instruction encoders improve performance, and can  use more instructions effectively. These results demonstrate the potential of language for compositionally representing large spaces of complex plans.

\subsection{Qualitative Analysis}
Observing games played by our model, we find that most instructions are both generated and executed as humans plausibly would. The \emph{executor} is often able to correctly count the number of units it should create in commands such as \emph{build 3 dragons}. 

There are several limitations. The \emph{executor} sometimes acts without instructions---partly due to mimicking some humans behaviour, but also indicating a failure to learn dependencies between instructions and actions. The \emph{instructor} sometimes issues commands which are impossible in its state (e.g. to attack with a unit that the it does not have)---causing weak behaviour from \emph{executor} model. 

\section{Related work}

Previous work has used language to specify exponentially many policies ~\citep{oh2017zero,andreas2017,yarats2018}, allowing zero-shot generalization across tasks. 
We develop this work by generating instructions as well as executing them. We also show how complex tasks can be decomposed into a series of instructions and executions.

Executing natural language instructions has seen much attention. The task of grounding language into an executable representation is sometimes called semantic parsing \citep{zettlemoyer:2007}, and has been applied to navigational instruction following, e.g. \citep{artzi:2013}.
More recently, neural models instruction following have been developed for a variety of domains, for example \citep{mei2016listen} and \citep{hermann2017grounded}.
Our dataset offers a challenging new problem for instruction following, as different instructions will apply to different subsets of available units, and multiple instructions may be apply at a given time.


Instruction generation has been studied as a separate task. \citep{daniele2017navigational} map navigational paths onto instructions. 
\citep{fried2017unified} generate instructions for complex tasks that humans can follow, and
\citep{fried2018speaker} train a model for instruction generation, which is used both for data augmentation and for pragmatic inference when following human-generated instructions. We build on this work by also generating instructions at test time, and showing that latent language improves performance.

Learning to play a complete real-time strategy game, including unit building, resources gathering, defence, invasion, scouting, and expansion, remains a challenging problem ~\citep{ontanon2013survey}, in particular due to the complexity and variations of commercially successful games (e.g., StarCraft I/II), and its demand of computational resources. Traditional approaches focus on sub-tasks with hand-crafted features and value functions (e.g., building orders~\citep{churchill2011build},  spatial placement of building~\citep{certicky2013implementing}, attack tactics between two groups of units~\citep{churchill2012fast}, etc). Inspired by the recent success of deep reinforcement learning, more works focus on training a neural network to finish sub-tasks~\citep{usunier2016episodic,bicnet}, some with strong computational requirement~\citep{zambaldi2018relational}. For full games, ~\citep{tian2017elf} shows that it is possible to train an end-to-end agent on a small-scaled RTS game with predefined macro actions, and TStarBot~\citep{sun2018tstarbots} applies this idea to StarCraft II and shows that the resulting agent can beat carefully-designed, and even cheating rule-based AI. By using human demonstrations, we hand crafting macro-actions. 

Learning an end-to-end agent that plays RTS games with unit-level actions is even harder. Progress is reported for MOBA games, a sub-genre of RTS games with fewer units---for example, ~\citep{openaifive} shows that achieving professional level of playing DoTA2 is possible with massive computation, 
and ~\citep{wu2018hierarchical} shows that with supervised pre-training on unit actions, and hierarchical macro strategies, a learned agent on Honor of Kings is on par with a top 1\% human player. 


\section{Conclusion}
We introduced a framework for decomposing complex tasks into steps of planning and execution, connected with a natural language interface.
We experimented with this approach on a new strategy game which is simple to learn but features challenging strategic decision making. 
We collected a large dataset of human instruction generations and executions, and trained models to imitate each role.
Results show that exploiting the compositional structure of natural language improves generalization for both the \emph{instructor} and \emph{executor} model, significantly outperforming agents without latent language.
Future work should use reinforcement learning to further improve the planning and execution models, and explore generating novel instructions.

\bibliography{main}
\bibliographystyle{plain}
\newpage

\section*{Appendix}

\appendix

\section{Detailed game design}
We develop an RTS game based on the MiniRTS framework, aspiring to make it intuitive for humans, while still providing a significant challenge to machines due to extremely high-dimensional observation and actions spaces, partial observability, and non-stationary environment dynamics imposed by the opponent. Below we describe the key game concepts.

\subsection{Game units specifications}  
\paragraph{Building units} Our game supports 6 different building types, each implementing a particular function in the game. Any building unit can be constructed by the \textsc{Peasant} unit type at any available map location by spending a specified amount of resources. Later, the constructed building can be used to construct units. Most of the building types can produce up to one different unit type, except of \textsc{Workshop}, which can produce 3 different unit types. This property of the \textsc{Workshop} building allows various strategies involving bluffing. A full list of available building units can be found in~\cref{table:buildings}.

\paragraph{Army units}

\begin{wrapfigure}{rh!}{0.48\linewidth}

\centering
\includegraphics[width=0.7\linewidth]{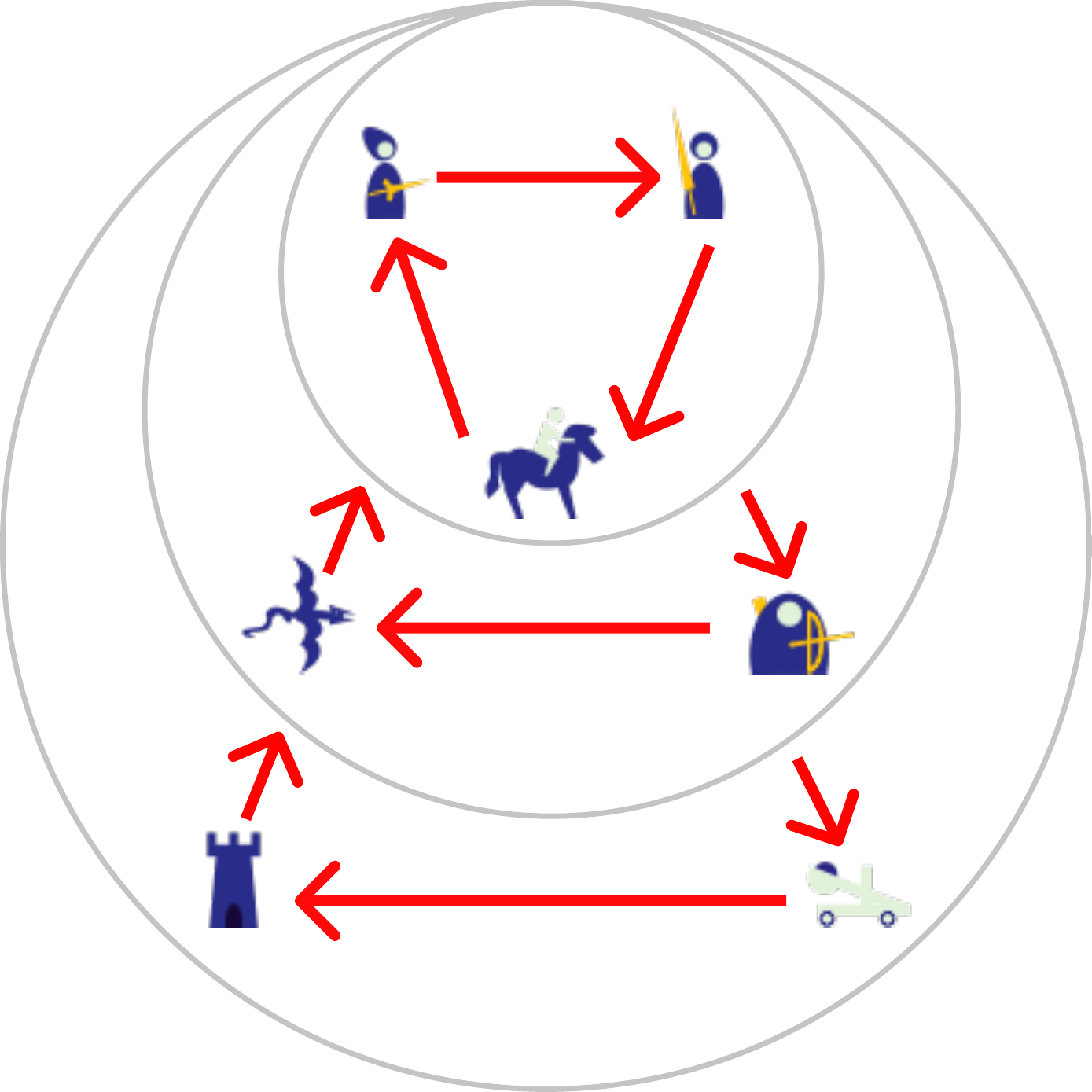}

\caption{\label{figure:attack_graph}
Our game implements the \emph{rock-paper-scissors} attack graph, where each unit has some units it is effective against and vulnerable to.
}
\end{wrapfigure}

The game provides a player with 7 army unit types, each having different strengths and weaknesses. \textsc{Peasant} is the only unit type that can construct building units and mine resources, so it is essential for advancing to the later stages of the game. We design the attack relationships between each unit type with a \emph{rock-paper-scissors} dynamic---meaning that each unit type has another unit type that it is effective against or vulnerable to. This property means that effective agents must be reactive to their opponent's strategy. See~\cref{figure:attack_graph} for a visualization. Descriptions of army units can be found in~\cref{table:units}.

\begin{table}[b]
\centering
\begin{tabular}{|p{.18\columnwidth}|p{.76\columnwidth}|}
\hline
Building name         & Description      \\
\hline
\textsc{Town Hall} & \begin{tabular}{@{}l@{}} The main building of the game, it allows a player to train \textsc{Peasant}s and \\serves as a storage for mined resources. \end{tabular}\\
\textsc{Barrack} & Produces \textsc{Spearmen}. \\
\textsc{Blacksmith} & Produces \textsc{Swordmen}. \\
\textsc{Stable} & Produces \textsc{Cavalry}. \\
\textsc{Workshop} & \begin{tabular}{@{}l@{}}  Produces \textsc{Catapult}, \textsc{Dragon} and \textsc{Archer}. The only building that can \\produce multiple unit types. \end{tabular}\\
\textsc{Guard Tower} & A building that can attack enemies, but cannot move.\\
\hline
\end{tabular}\\
\caption{\label{table:buildings} The list of the building units available in the game.}
\end{table}

\paragraph{Resource unit}
\textsc{Resource} is a stationary and neutral unit type, it cannot be constructed by anyone, and is only created during the map generation phase. \textsc{Peasant}s of both teams are allowed to mine the same \textsc{Resource} unit, until it is exhausted. Initial capacity is set to 500, and one mine action subtracts 10 points from the \textsc{Resource}. Several \textsc{Resource} units are placed randomly on the map, which gives raise to many strategies around \textsc{Resource} domination.

\subsection{Game map}
We represent the game map as a discrete grid of 32x32. Each cell of the grid can either be grass or water, where the grass cell is passable for any army units, while the water cell prevents all units except of \textsc{Dragon} to go through. Having water cells around one's main base can be leveraged as a natural protection. We generate maps randomly for each new game, we first place  one \textsc{Town Hall} for each player randomly. We then add some water cells onto the map, making sure that there is at least one path between two opposing \textsc{Town Hall}s, but otherwise aiming to create bottlenecks. Finally, we randomly locate several \textsc{Resource} units onto the map such that they are approximately equidistant from the players \textsc{Town Hall}s.

\begin{table}[h!]
\centering
\begin{tabular}{|p{.18\columnwidth}|p{.76\columnwidth}|}
\hline
Unit name         & Description      \\
\hline
\textsc{Peasant} & Gathers minerals and constructs buildings, not good at fighting. \\
\textsc{Spearman} & Effective against cavalry. \\
\textsc{Swordman} & Effective against spearmen. \\
\textsc{Cavalry} & Effective gainst swordmen. \\
\textsc{Dragon} & Can fly over obstacles, can only be attacked by archers and towers. \\
\textsc{Archer} & Great counter unit against dragons. \\
\textsc{Catapult} & Easily demolishes buildings. \\

\hline
\end{tabular}\\
\caption{\label{table:units} The list of the army units available in the game.}
\end{table}

\section{RTS game as an Reinforcement Learning environment} 
Our platform can be also used as an RL environment. In our code base we implement a framework that allows a straightforward interaction with the game environment in a canonical RL training loop. Below we detail the environment properties.

\subsection{Observation space}
We leverage both spatial representation of the map, as well as internal state of the game engine (e.g. units health points and attacking cool downs, the amount of resources, etc.) to construct an observation. We carefully address the fog of war, by masking out the regions of the map that have not been visited. In addition, we remove any unseen enemy units attributes from the observation. The partial observability of the environment makes it especially challenging to apply RL due to highly non-stationary state distribution.

\subsection{Action space}
At each timestep of the environment we predict an action for each of our units, both buildings and army. The action space is consequently large---for example, any unit can go to any location at each timestep. Prediction of an unit action proceeds in steps, we first predict an action type (e.g. \textsc{Move} or \textsc{Attack}), then, based on the action type, we predict the action outputs. For example, for the \textsc{Build Building} action type the outputs will be the type of the future building and its location on the game map. We summarize all available action types and their structure in~\cref{table:actions}.

\begin{table}[t]
\centering

\begin{tabular}{|p{.20\columnwidth}|p{.2\columnwidth}|p{.5\columnwidth}|}
\hline
Action Type & Action Output & Input Features \\
\hline
\textsc{Idle}          & \texttt{NULL}    &   \texttt{NULL} \\
\textsc{Continue}      & \texttt{NULL}     &  \texttt{NULL} \\
\textsc{Gather}        & \texttt{resource\_id}    & \texttt{resources\_features} \\
\textsc{Attack}        & \texttt{enemy\_unit\_id}     & \texttt{enemy\_units\_features} \\
\textsc{Train Unit}   & \texttt{unit\_type}     & \texttt{unit\_type\_features} \\
\textsc{Build Building}& \texttt{unit\_type}, \texttt{(x,y)} & \texttt{unit\_type\_features}, \texttt{map\_cells\_features}  \\
\textsc{Move}          & \texttt{(x,y)}     & \texttt{map\_cells\_features}  \\
\hline
\end{tabular}
\caption{\label{table:actions}  We implement a separate action classifier per action type, because each action type needs to model a probability distribution over different objects (Action Output). For example, for the \textsc{Attack} action we need estimate a probability distribution over all visible enemy units and predict an enemy unit id, or \textsc{Build Building} action needs to model two probability distributions, one over building type to be constructed, and another over all possible $(x,y)$ discrete location on the map where the future building will be placed.}
\end{table}

\subsection{Reward structure}
We support a sparse reward structure, e.g. the reward of 1 is issued to an agent at the end if the game is won, all the other timesteps receive the reward of 0. Such reward structure makes exploration an especially challenging given the large dimensionality of the action space and the planning horizon.

\section{Data collection}

We design a data collection task based on ParlAI, a transparent framework to interact with human workers. We develop separate game control interfaces for both the \emph{instructor} and the \emph{executor} players, and ask two humans to play the game collaboratively against a rule-based AI opponent. Both player have the same access to the game observation, but different control abilities.

The \emph{instructor} control interface allows the human player to perform the following actions:
\begin{itemize}
    \item \textbf{Issue} a natural language instruction to the \emph{executor} at any time of the game. We allow any free-form language instruction. 
    \item \textbf{Pause} the game flow at any time. Pausing allows the \emph{instructor} to analyze the game state more thoroughly and plan strategically.
    \item \textbf{Warn} the \emph{executor} player in case they do not follow issued instructions precisely. This option allows us to improve data quality, by filtering \emph{executor}s who do not follow instructions. 
\end{itemize}
On the other hand, the \emph{executor} player gets to:
\begin{itemize}
    \item \textbf{Control} the game units by direct manipulation using computer's input devices (e.g. mouse). The \emph{executor} is tasked to complete the current instruction, rather than to win the game.
    \item \textbf{Ask} the \emph{instructor} for either a new instruction, or a clarification. 
\end{itemize}

Each human workers is assigned with either the \emph{instructor} or the \emph{executor} role randomly, thus the same person can experience the game on both ends over multiple attempts.

\begin{table}[t]
\centering
\begin{tabular}{|p{.2\columnwidth}|p{.74\columnwidth}|}
\hline
Strategy Name         & Description      \\
\hline
\textsc{Simple}                        &  This strategy first sends all 3 initially available \textsc{Peasant}s to mine to the closest resource, then it chooses one army unit type from \textsc{Spearman}, \textsc{Swordman}, \textsc{Cavalry}, \textsc{Archer}, or \textsc{Dragon}, then it constructs a corresponding building, and finally trains 3 units of the selected type and sends them to attack. The strategy then continuously maintains the army size of 3, in case an army unit dies.    \\
\textsc{Medium}                        &  Same as \textsc{Simple} strategy, only the size of the army is randomly selected between 3 and 7.  \\
\textsc{Strong}                        &  This strategy is adaptive, and it reacts to the opponent's army. In particular, this strategy constantly scouts the map using one \textsc{Peasant} and to lean the opponent's behaviour. Once it sees the opponent's army it immediately trains a counter army based on the attack graph (see~\cref{figure:attack_graph}). Then it clones the \textsc{Medium} strategy.  \\
\textsc{Second Base}                        &  This strategy aims to build a second \textsc{Town Hall} near the second closest resource and then it uses the double income to build a large army of a particular unit type. The other behaviours is the same as in the \textsc{Medium} strategy.  \\
\textsc{Tower Rush}                        &  A non-standard strategy, that first scouts the map in order to find the opponent using a spare \textsc{Peasant}. Once it finds it, it starts building \textsc{Guard Tower}s close to the opponent's \textsc{Town Hall} so they can attack the opponent's units.    \\
\textsc{Peasant Rush}                        &  This strategy sends first 3 \textsc{Peasant}s to mine, then it keeps producing more \textsc{Peasant}s and sending them to attack the opponent. The hope of this strategy is to beat the opponent by surprise.  \\
\hline
\end{tabular}\\
\caption{\label{table:rule_based_bots} The rule-based strategies we use as an opponent to the human players during data collection.}
\end{table}

\subsection{Quality control}
To make sure that we collect data of high quality we take the following steps:
\paragraph{Game manual} We provide a detailed list of instructions to a human worker at the beginning of each game and during the game's duration. This manual aims to narrate a comprehensive overview various game elements, such as player roles, army and building units, control mechanics, etc. We also record several game replays that serve as an introductory guideline to the players.
\paragraph{Onboarding} We implement an onboarding process to make sure that novice players are comfortable with the game mechanics, so that they can play  with other players effectively. For this, we ask a novice player to perform the \emph{executor}'s duties and pair them with a bot that issues a pre-defined set of natural language instructions that implements a simple walkthrough strategy. We allocate enough time for the human player to work on the current instruction, and to also get comfortable with the game flow. We let the novice player play several games until we verify that they pass the required quality bar. We assess the performance of the player by running a set of pattern-matching scripts that verify if the performed control actions correspond to the issued instructions (for example, if an instruction says "build a barrack", we make sure that the player executes the corresponding low-level action). If the human player doesn't pass our qualification requirements within 5 games, we prevent them from participating in our data collection going forward and filter their games from the dataset. 

\paragraph{Player profile} We track performance of each player, breaking it down by a particular role (e.g. \emph{instructor} or \emph{executor}). We gather various statistics about each player and build a comprehensive player profile. For example, for the \emph{instructor} role we gather data such as overall win rate, the number of instructions issued per game, diversity of issued instructions; for the \emph{executor} role we monitor how well they perform on the issued instruction (using a pattern matching algorithm), the number of warnings they receive from the \emph{instructor}, and many more. We then use this profile to decide whether to upgrade a particular player to playing against stronger opponents (see ~\cref{rule_based_ai}) in case they are performing well, or prevent them from participating in our data collection at all otherwise.
\paragraph{Feedback}
We use several initial round of data collection as a source of feedback from the human players. The received feedback helps us to improve the game quality. Importantly, after we finalize the game configuration, we disregard all the previously collected data in our final dataset.

\paragraph{Final filtering} Lastly, we take another filtering pass against all the collected game replays and eliminate those replays that don't meet the following requirements:
\begin{itemize}
    
    \item A game should have at least $3$ natural language instructions issued by the \emph{instructor}.
    \item A game should have at least $25$ low-level control actions issued by the \emph{executor}.
\end{itemize}

By implementing all the aforementioned safe guards we are able to gather a high quality dataset. 

\subsection{Rule-based bots}
\label{rule_based_ai}
We design a set of diverse game strategies that are implemented by our rule-based bots (~\cref{table:rule_based_bots}). Our handcrafted strategies explore much of the possibilities that the game can offer, which in turn allows us to gather a multitude of emergent human behaviours in our dataset. Additionally, we employ a resource scaling hyperparameter, which controls the amount of resources a bot gets during mining. This hypermarameter offers a finer control over the bot's strength, which we find beneficial for onboarding novice human players. 
We pair a team of two human players (the \emph{instructor} and \emph{executor}) with a randomly sampled instance of a rule-based strategy and the resource scaling hyperparameter during our data collection, so the human player doesn't know in advance who is their opponent. This property rewards reactive players. We later observe that our models are able to learn the scouting mechanics from the data, which is a crucial skill to be successful in our game.

\begin{figure*}[t!]

\begin{subfigure}[t]{0.48\textwidth}
\centering
\includegraphics[width=0.97\textwidth]{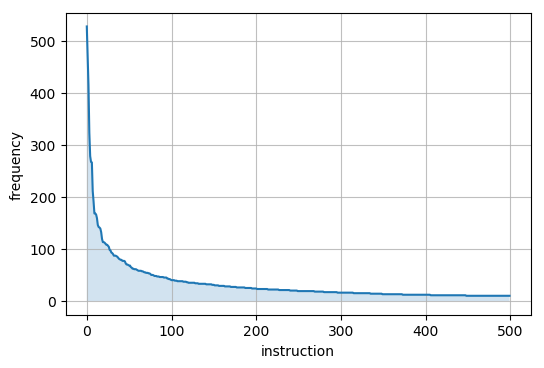}
\caption{Top $500$ most frequent instructions }

\end{subfigure}
 ~
\begin{subfigure}[t]{0.48\textwidth}
\centering
\includegraphics[width=\textwidth]{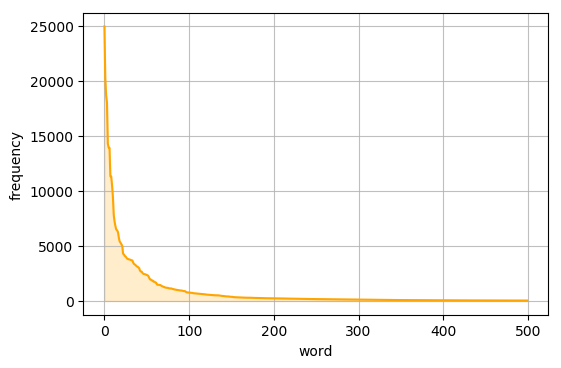}
\caption{Top $500$ most frequent words }

\end{subfigure}

\caption{Frequency histograms for the dataset instructions and words.}
    \label{fig:historgram}
\end{figure*}

\section{Model architecture}

\subsection{Convolutional channels of Spatial Encoder}
We use the following set of convolutional channels to extract different bits of information from spatial representation of the current observation.
\begin{enumerate}
    \item \textbf{Visibility}: 3 binary channels for each state of visibility of a cell (\textsc{Visible}, \textsc{Seen}, and \textsc{Invisible}).
    \item \textbf{Terrain}: 2 binary channels for each terrain type of a cell (grass or water).
    \item \textbf{Our Units}: 13 channels for each unit type of our units. Here, a cell contains the number  of our units of the same type located in it.
    \item \textbf{Enemy Units}: similarly 13 channels for visible enemy units.
    \item \textbf{Resources}: 1 channel for resource units.
\end{enumerate}

\subsection{Action Classifiers}
\begin{figure*}[b!]

\centering
\includegraphics[width=0.7\linewidth]{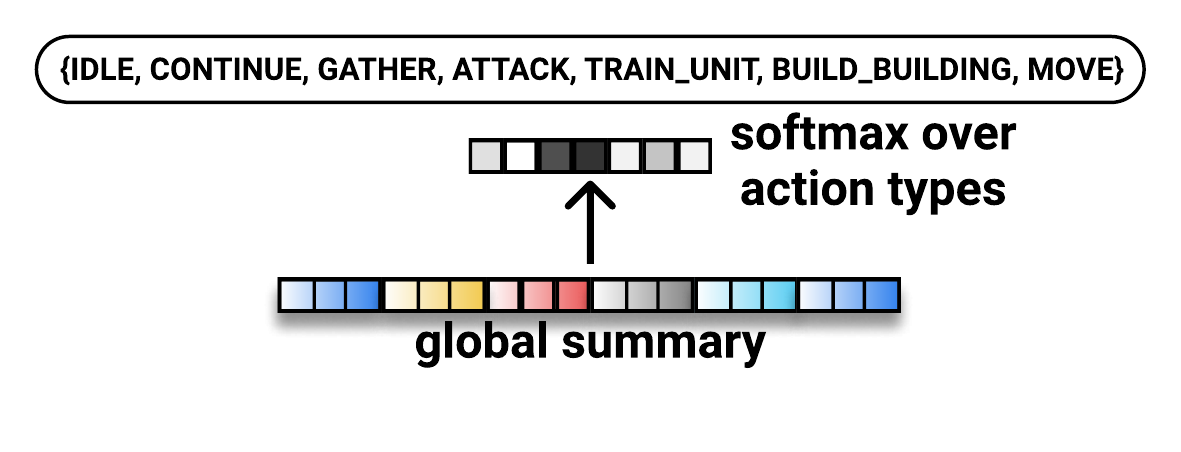}

\caption{\label{figure:act_type}
The \textsc{Action Type} classifier is parameterized as an MLP network to model a softmax distribution over action types based on the unit's global summary features vector.
}
\end{figure*}

\begin{figure*}[t!]

\centering
\includegraphics[width=0.4\linewidth]{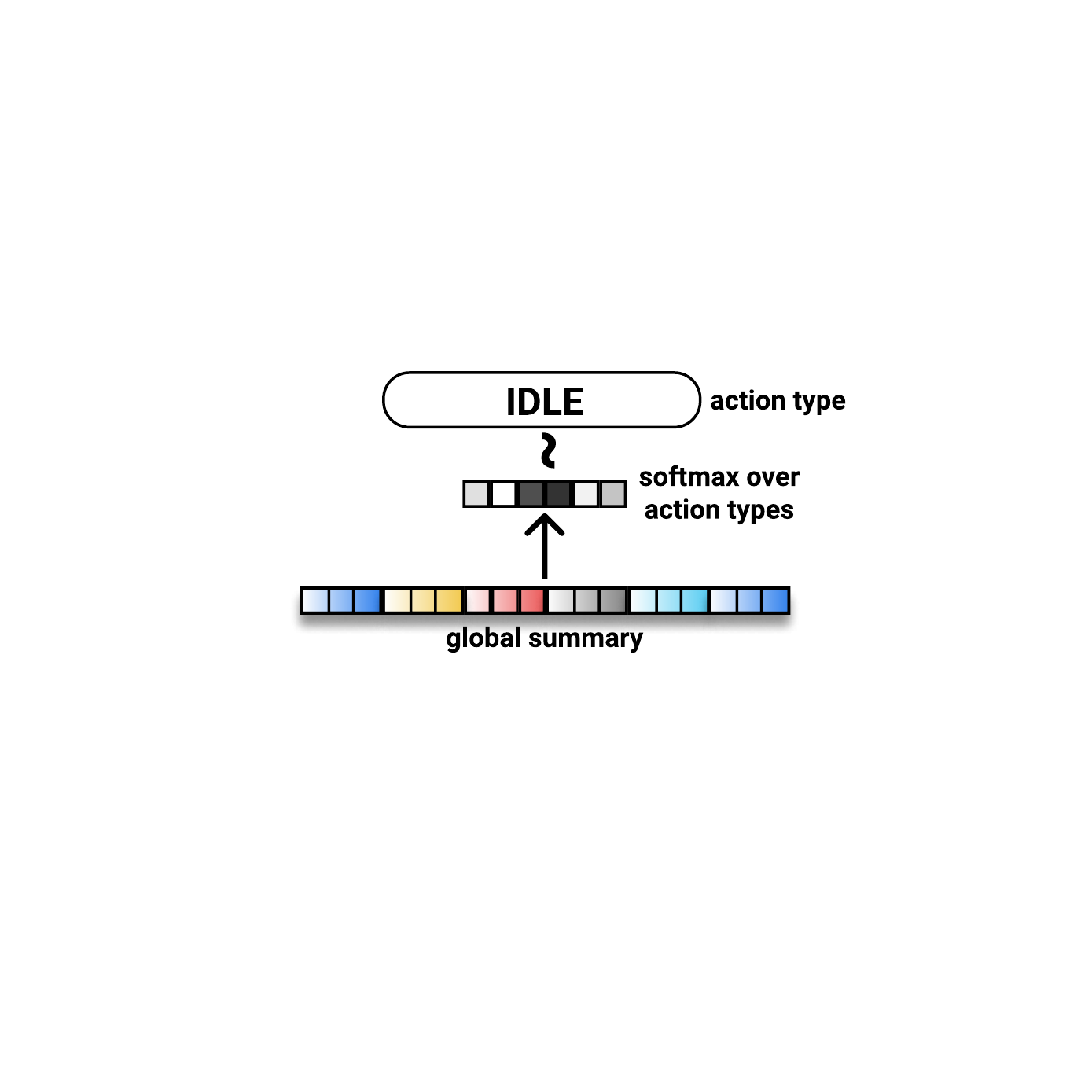}
\includegraphics[width=0.4\linewidth]{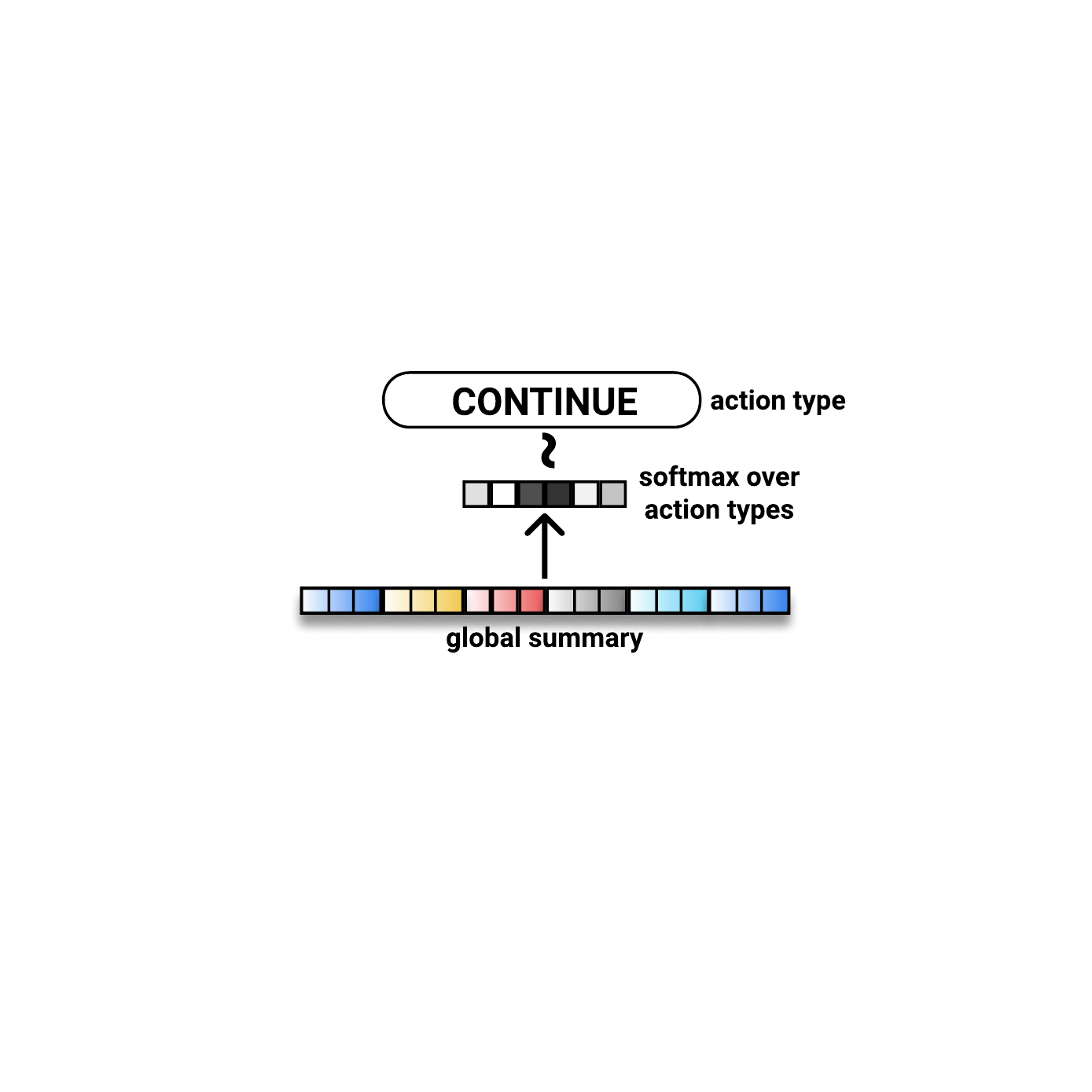}
\includegraphics[width=0.4\linewidth]{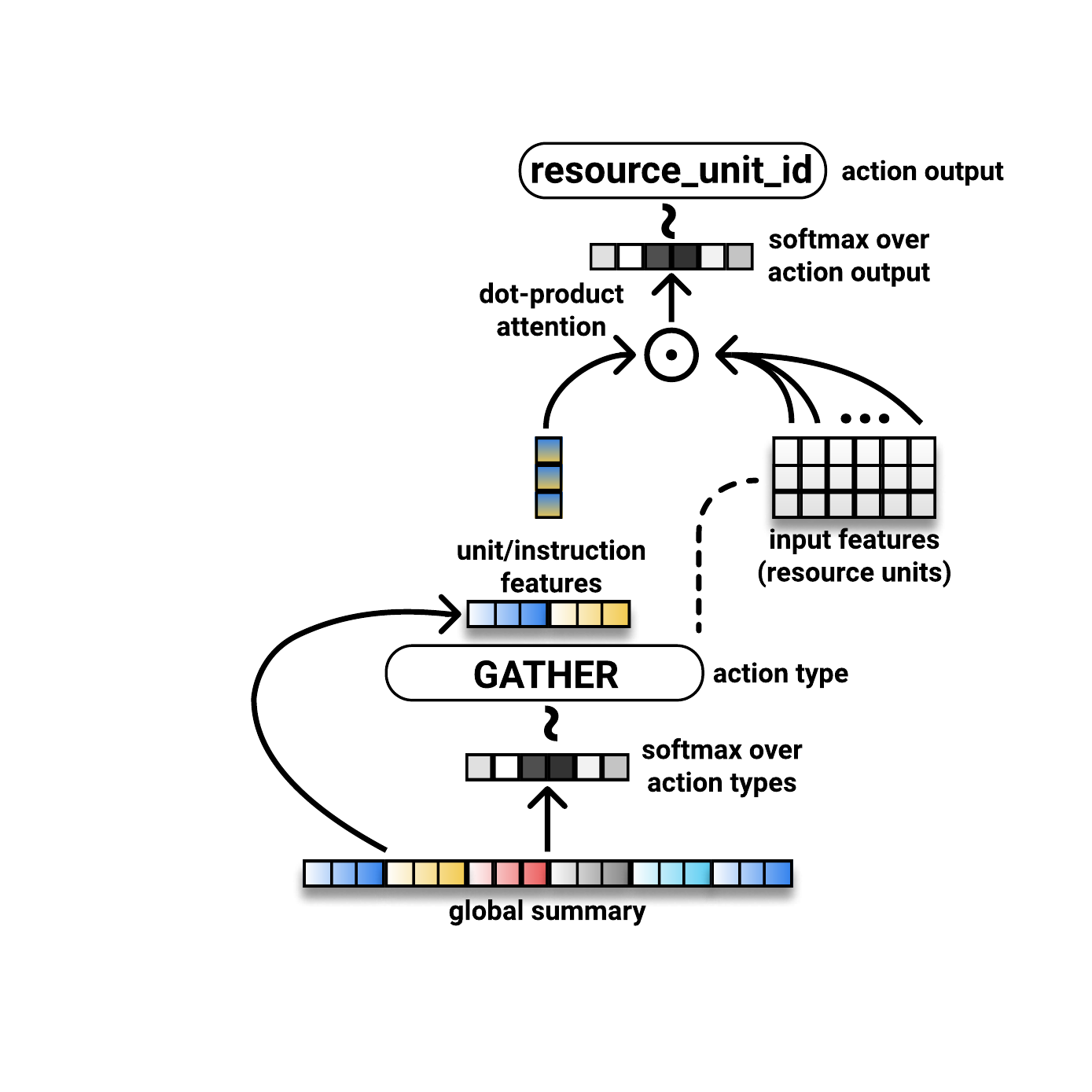}
\includegraphics[width=0.4\linewidth]{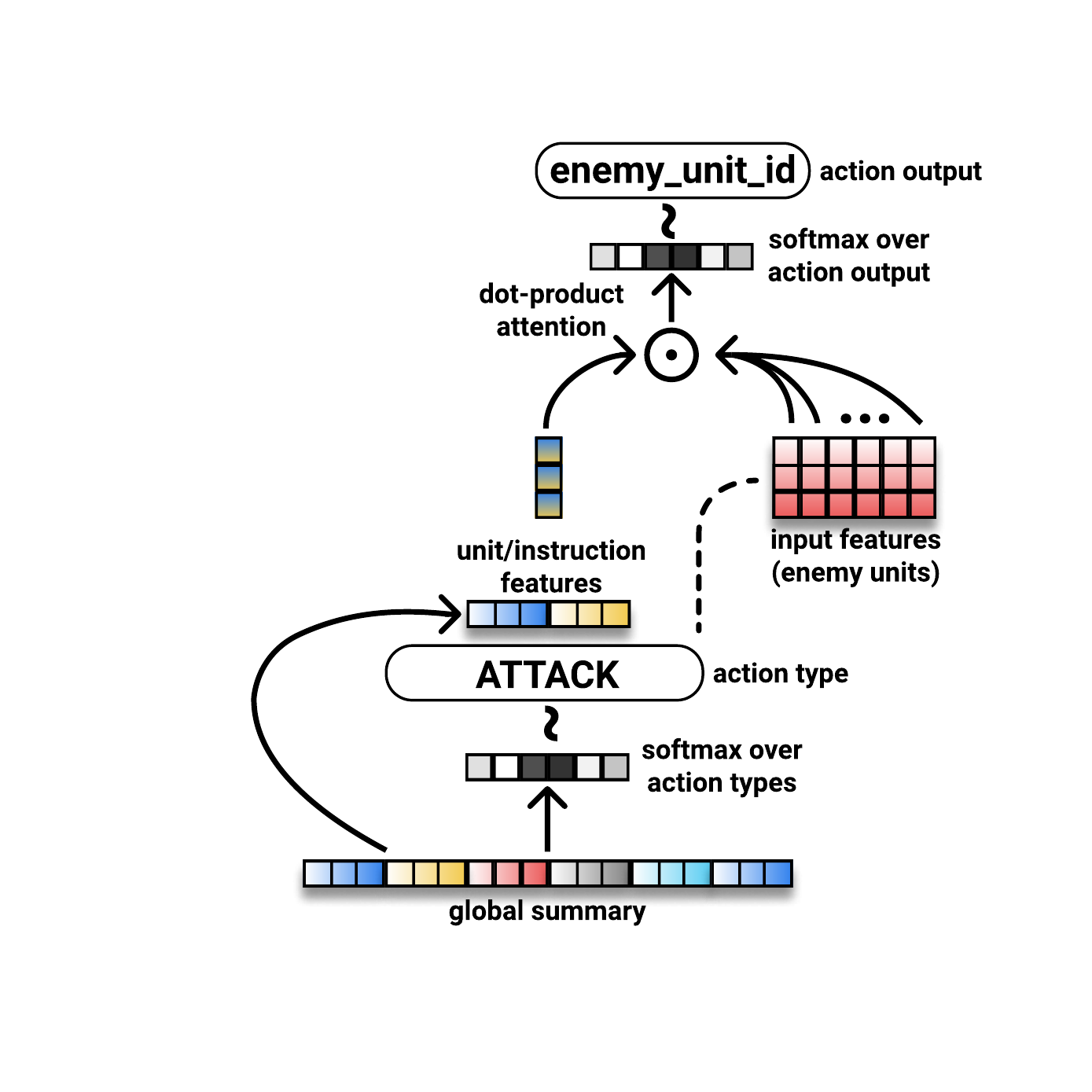}

\includegraphics[width=0.4\linewidth]{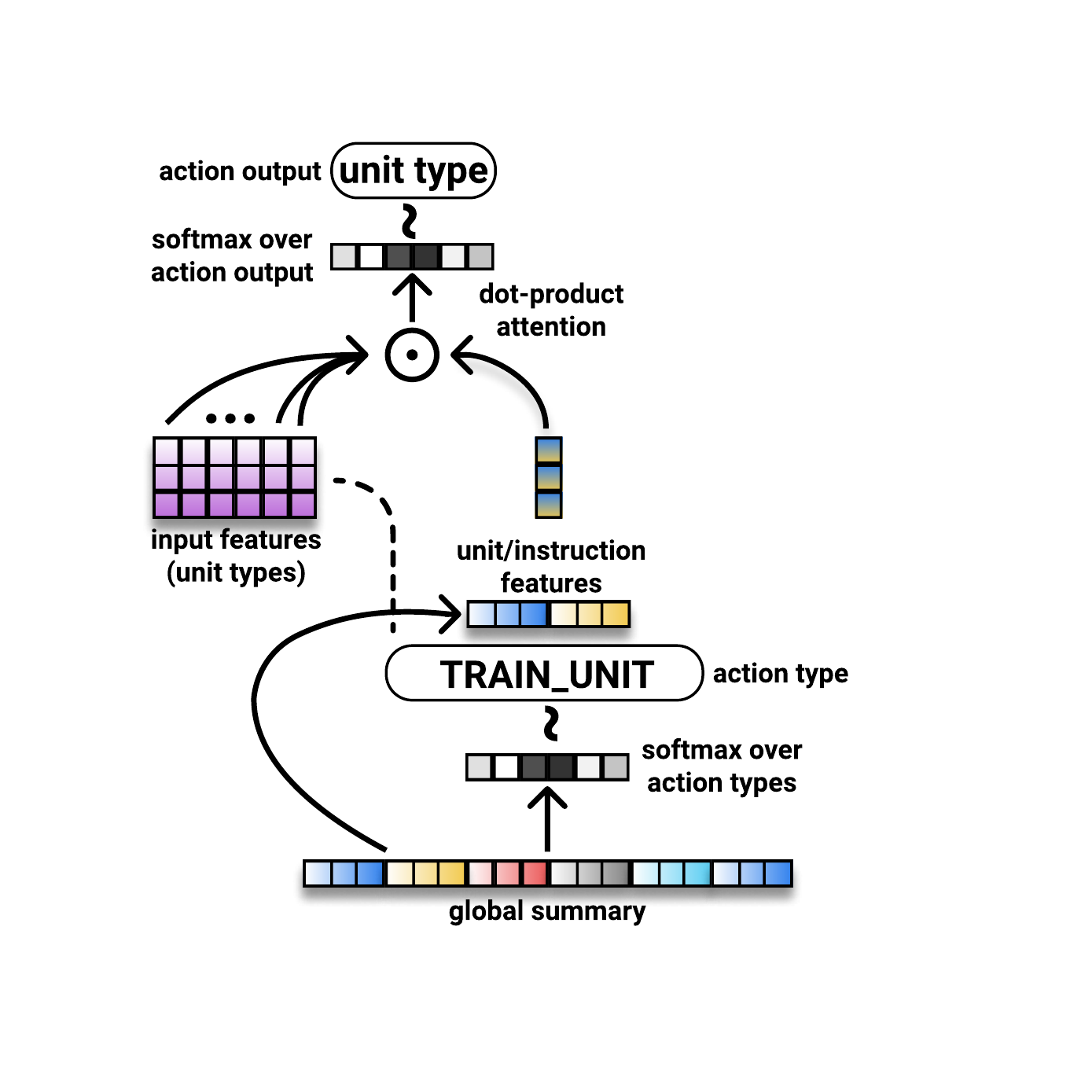}
\includegraphics[width=0.4\linewidth]{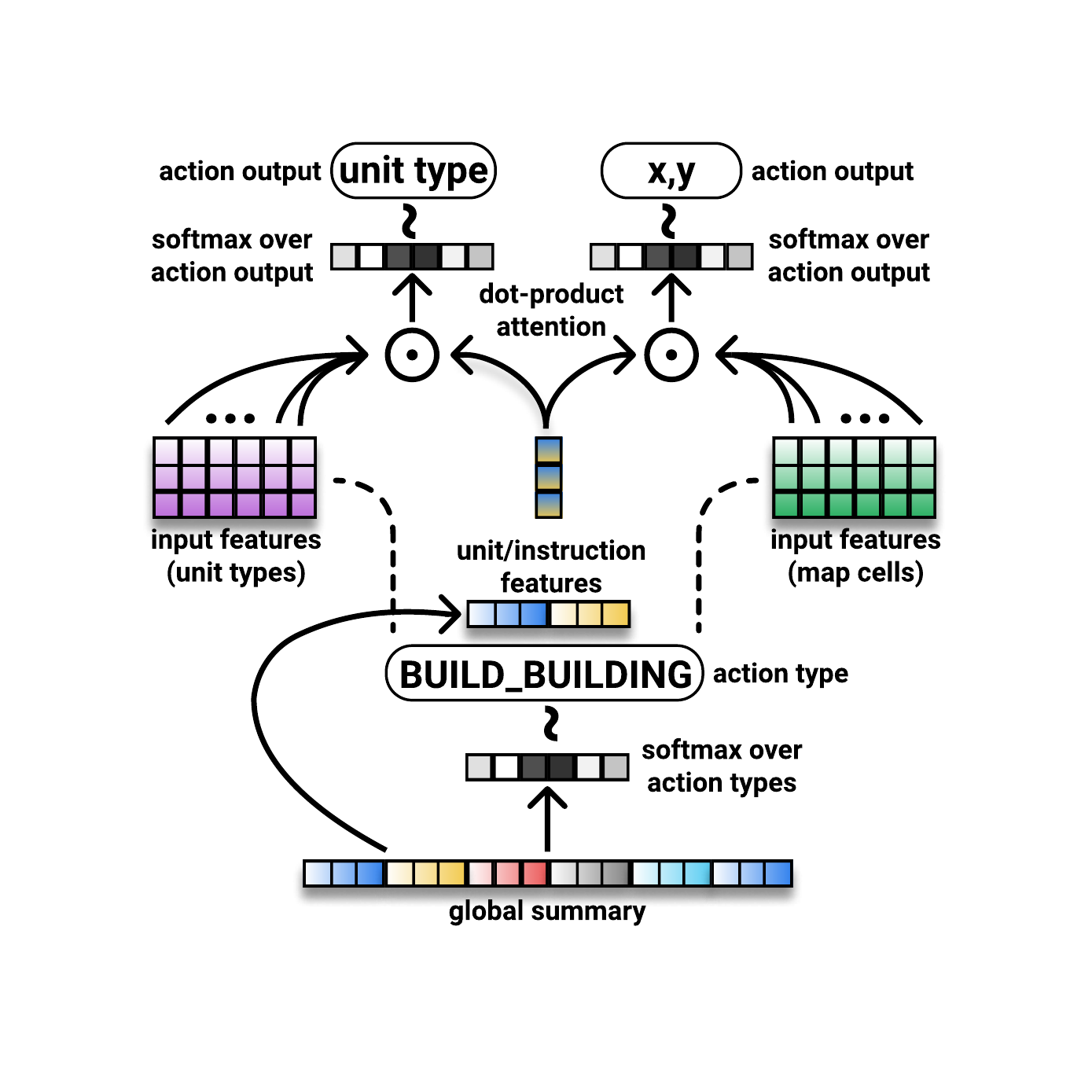}
\includegraphics[width=0.4\linewidth]{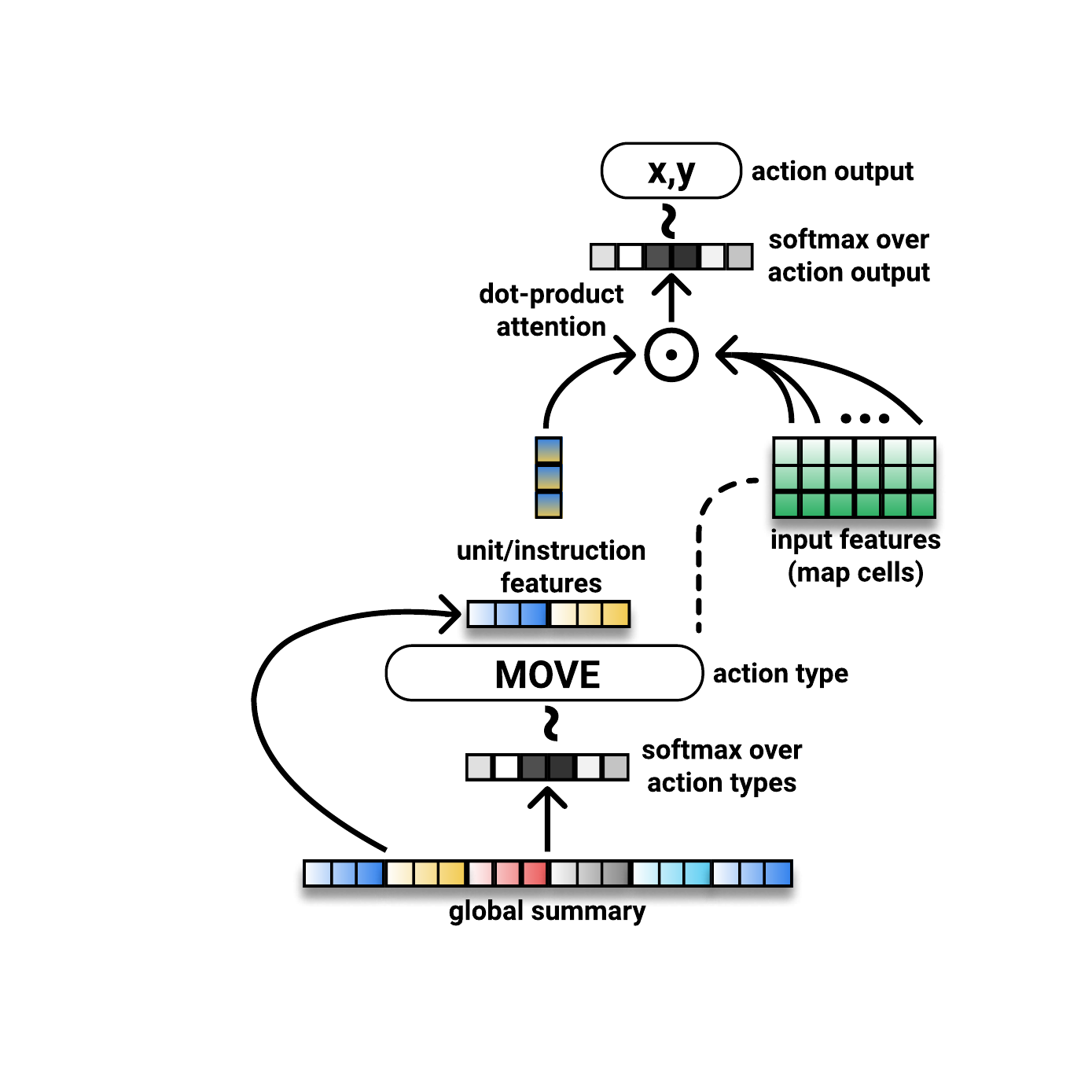}

\caption{\label{figure:classifiers_graphs}
Separate classifiers for each of the available action types.
}
\end{figure*}

At each step of the game we predict actions for each of the player's units, we do this by performing a separate forward pass for ofv the following network for each unit. Firstly, we run an MLP (\cref{figure:act_type}) based action classifier to sample the unit's \textsc{Action Type}. We feed the unit's global summary features (see Fig. 3 of the main paper) into the classifier and sample an action type (see~\cref{table:actions} for the full list of possible actions). Then, given the sampled action type we predict the \textsc{Action Output} based on the unit's features, unit dependent instructions features, and the action input features. We provide an overview of \textsc{Action Output}s and \textsc{Input Features} for each actions in \cref{table:actions}. In addition, you can refer to the diagram~\cref{figure:classifiers_graphs}.

\section{Dataset details}
Through our data collection we gather a dataset of over 76 thousand of instructions and corresponding executions.  We observe a wide variety of different strategies and their realizations in natural language. For example, we observe emergence of complicated linguistic constructions (\cref{table:instruction_examples}).

We also study the distribution of collected instructions. While we notice that some instructions are more frequent than others, we still observe a good coverage of strategies realizations, which serve as a ground for generalization. In~\cref{table:instr_freq} we provide a list of most frequently used instructions, and in~\cref{fig:historgram} shows the overall frequency distribution for instructions and words in our dataset.

Finally, we provide a random sample of 50 instructions from our dataset in~\cref{table:sample_instructions}, where showing the diversity and complexity of the collected instructions.

\begin{table}[t!]
\centering
\begin{tabular}{|p{.4\columnwidth}|p{.54\columnwidth}|}
\hline
Linguistic Phenomena         & Example      \\
\hline
Counting                        &  \emph{Build 3 dragons.}  \\
Spatial Reference                       &  \emph{Send him to the choke point behind the tower.}  \\
Locations                       &  \emph{Build one to the left of that tower.}  \\
Composed Actions                &  \emph{Attack archers, then peasants.}  \\
Cross-instruction anaphora      &  \emph{Use it as a lure to kill them.}  \\
\hline
\end{tabular}
\caption{\label{table:instruction_examples} Complex linguistic phenomena emerge as humans instruct others how to play the game.}
\end{table}

\begin{table}[ht]
\centering
\begin{tabular}{|p{.96\columnwidth}|}
\hline
Instruction     \\
\hline
\emph{Build 1 more cavalry}.\\
\emph{Attack peaons}.\\
\emph{Build barrack in between south pass at new town}.\\
\emph{Have all peasants gather minerals next to town hall}.\\
\emph{Have all peasants mine ore}.\\
\emph{Fight u peaas}.\\
\emph{Stop the peasants from mining}.\\
\emph{Build a new town hall between the two west minerals patches}.\\
\emph{Build 2 more swords}.\\
\emph{Use cavalry to attack enemey}.\\
\emph{Explore and find miners}.\\
\emph{If you see any idle peasants  please have them build}.\\
\emph{Okay that doesn't work then  build them on your side of the wall then}.\\
\emph{Create 4 more archers}.\\
\emph{Make a new town hall in the middle of all 3}.\\
\emph{Attack tower with catas}.\\
\emph{Kill cavalry and peasants then their townhall}.\\
\emph{Attack enemy peasants with cavalry as well}.\\
\emph{Send all peasants to collect minerals}.\\
\emph{Attack enemy peasant}.\\
\emph{Keep creating peasants and sending them to mine}.\\
\emph{Send one catapult to attack the northern guard tower send a dragon for protection}.\\
\emph{Send all but 1 peasant to mine}.\\
\emph{Mine with the three peasants}.\\
\emph{Use that one to scout and don't stop}.\\
\emph{Bring scout back to base to mine}.\\
\emph{You'll need to attack them with more peasants to kill them}.\\
\emph{Build a barracks}.\\
\emph{Send all peasants to find a mine and mine it}.\\
\emph{Start mining there with your 3}.\\
\emph{Make four peasants}.\\
\emph{Move archers west then north}.\\
\emph{Attack with cavalry}.\\
\emph{Make two more workers}.\\
\emph{Make 2 more calvary and send them over with the other ones}.\\
\emph{Return to base with scout}.\\
\emph{Build 2 peasants at the new mine}.\\
\emph{If attacked  retreat south}.\\
\emph{Make the rest gather minerals too}.\\
\emph{All peasants flee the enemy}.\\
\emph{Attack the peasants in the area}.\\
\emph{Attack the last archer with all peasants on the map}.\\
\hline
\end{tabular}
\caption{\label{table:sample_instructions} Examples of randomly sampled instructions.}
\end{table}

\newpage

\begin{table}[t]
\centering
\begin{tabular}{|p{.31\columnwidth}|p{.12\columnwidth}||p{.31\columnwidth}|p{.12\columnwidth}|}
\hline
Instruction         & Frequency  & Instruction & Frequency    \\
\hline
\emph{Attack}. & 527 & \emph{Send idle peasants to mine}. & 68 \\
\emph{Send all peasants to mine}. & 471 & \emph{Attack that peasant}. & 68 \\
\emph{Build a workshop}. & 414 & \emph{Send all peasants to mine minerals}. & 65 \\
\emph{Retreat}. & 323 & \emph{Build a barracks}. & 64 \\
\emph{Build a stable}. & 278 & \emph{Build barrack}. & 62 \\
\emph{Send peasants to mine}. & 267 & \emph{Return to mine}. & 62 \\
\emph{All peasants mine}. & 266 & \emph{Build peasant}. & 61 \\
\emph{Send idle peasant to mine}. & 211 & \emph{Build catapult}. & 61 \\
\emph{Build workshop}. & 191 & \emph{Create a dragon}. & 61 \\
\emph{Build a dragon}. & 168 & \emph{Mine with peasants}. & 60 \\
\emph{Kill peasants}. & 168 & \emph{Build 3 peasants}. & 59 \\
\emph{Attack enemy}. & 166 & \emph{Defend}. & 58 \\
\emph{Attack peasants}. & 159 & \emph{Build cavalry}. & 58 \\
\emph{Build a guard tower}. & 146 & \emph{Make an archer}. & 58 \\
\emph{Attack the enemy}. & 142 & \emph{Attack dragon}. & 58 \\
\emph{Stop}. & 141 & \emph{Send all peasants to collect minerals}. & 57 \\
\emph{Attack peasant}. & 139 & \emph{Defend base}. & 57 \\
\emph{Kill that peasant}. & 132 & \emph{Build 2 more peasants}. & 56 \\
\emph{Mine}. & 119 & \emph{Build 2 peasants}. & 55 \\
\emph{Build another dragon}. & 113 & \emph{Make 2 archers}. & 55 \\
\emph{Make another peasant}. & 113 & \emph{Make dragon}. & 54 \\
\emph{Build stable}. & 112 & \emph{Build 2 dragons}. & 54 \\
\emph{Make a dragon}. & 110 & \emph{Attack dragons}. & 54 \\
\emph{Build a blacksmith}. & 108 & \emph{Make a stable}. & 53 \\
\emph{Build a catapult}. & 108 & \emph{Make a catapult}. & 53 \\
\emph{Back to mining}. & 106 & \emph{Build 6 peasants}. & 52 \\
\emph{Build another peasant}. & 104 & \emph{Attack archers}. & 50 \\
\emph{Make a peasant}. & 98 & \emph{Kill all peasants}. & 50 \\
\emph{Build a barrack}. & 97 & \emph{Build 2 catapults}. & 50 \\
\emph{Build 4 peasants}. & 93 & \emph{Idle peasant mine}. & 49 \\
\emph{Have all peasants mine}. & 92 & \emph{Make peasant}. & 48 \\
\emph{Build 2 archers}. & 90 & \emph{Attack enemy peasant}. & 48 \\
\emph{Build dragon}. & 87 & \emph{Attack archer}. & 48 \\
\emph{Attack with peasants}. & 87 & \emph{Build another archer}. & 47 \\
\emph{Return to mining}. & 87 & \emph{Make 4 peasants}. & 47 \\
\emph{Build a peasant}. & 86 & \emph{Make 3 peasants}. & 47 \\
\emph{Idle peasant to mine}. & 85 & \emph{Build 2 more archers}. & 46 \\
\emph{Make a workshop}. & 83 & \emph{Send idle peasant back to mine}. & 46 \\
\emph{Create a workshop}. & 81 & \emph{Make more peasants}. & 46 \\
\emph{Mine with all peasants}. & 80 & \emph{Make 2 more peasants}. & 46 \\
\emph{Build 3 more peasants}. & 79 & \emph{Build blacksmith}. & 46 \\
\emph{Create another peasant}. & 79 & \emph{Collect minerals}. & 45 \\
\emph{Send all idle peasants to mine}. & 77 & \emph{Kill}. & 45 \\
\emph{Build 3 archers}. & 77 & \emph{Build an archer}. & 45 \\
\emph{Kill peasant}. & 77 & \emph{Keep mining}. & 45 \\
\emph{Make another dragon}. & 76 & \emph{Keep attacking}. & 43 \\
\emph{Kill him}. & 72 & \emph{Attack dragons with archers}. & 43 \\
\emph{Build guard tower}. & 70 & \emph{Create a stable}. & 42 \\
\emph{Attack town hall}. & 70 & \emph{Make 3 more peasants}. & 42 \\
\emph{Start mining}. & 69 & \emph{Attack the peasant}. & 41 \\

\hline
\end{tabular}
\caption{\label{table:instr_freq} The top 100 instructions sorted by their usage frequency.}
\end{table}

\newpage


\end{document}